\documentclass[10pt,twocolumn,letterpaper]{article}

\usepackage{cvpr}              
\definecolor{cvprblue}{rgb}{0.21,0.49,0.74}
\usepackage[pagebackref,breaklinks,colorlinks,allcolors=cvprblue]{hyperref}
\usepackage{booktabs}
\usepackage[table]{xcolor}
\def\confName{CVPR}
\def\confYear{2027}

\title{MMotion: Collaborative  Cross-Scale Memory for Text-to-Motion Generation}

\author{
Zhaoxin Fan\\
Beihang University\\
\and
Qingchen Yu\\
Beihang University
\and
Kejian Wu\\
XREAL
\and
Jian Yang\\
Beihang University
\and
Zeming Liu\\
Beihang University
\and
Wenjun Wu\\
Beihang University
\and
Lili Wang\\
Beihang University
}

\begin{document}
\maketitle
\begin{abstract}
Text-to-motion generation aims to synthesize realistic human motions from natural-language descriptions, enabling applications in animation, virtual avatars, and embodied agents. Despite recent progress, existing methods either retrieve motion exemplars or retain long-range context, but rarely exploit memory to jointly reason about semantics, dynamics, and generation history. In this paper, we propose MMotion, a collaborative multi-scale memory framework that unifies scale-conditioned semantic retrieval, latent dynamics modeling, and causal working memory. Specifically, semantic memory provides motion priors at different generation scales, dynamics memory captures latent state transitions, and working memory preserves causal context from previously generated motions. A collaborative fusion mechanism integrates these complementary memories to determine what motion to generate, how it should evolve, and how it should remain consistent with its history. MMotion is architecture-agnostic and can be readily integrated into both diffusion-based and next scale autoregressive models. Extensive experiments across standard benchmarks and diverse backbones demonstrate consistent improvements over strong baselines.
\end{abstract}    
\section{Introduction}
\label{sec:intro}

Text-to-motion generation aims to synthesize realistic human motion from natural-language descriptions. As a fundamental interface between language and human behavior, it has broad applications in character animation \cite{qing2023story}, virtual reality \cite{sahili2025text}, human–computer interaction \cite{park2025unified}, and embodied intelligence\cite{li2026language}. Beyond generating visually plausible poses, a successful model must accurately interpret compositional textual instructions, preserve fine-grained motion details \cite{khani2025motion}. Achieving these objectives simultaneously remains challenging because human motion exhibits complex dependencies across semantic, temporal, and skeletal scales.

Recent text-to-motion methods have progressed from deterministic sequence prediction to discrete motion modeling and probabilistic generation \cite{kim2025personabooth,zhou2026exges,fan2024everything2motion}. Autoregressive approaches \cite{guo2024momask,zheng2026next,jiang2023motiongpt} typically tokenize motion sequences and predict motion tokens conditioned on text, benefiting from the expressive sequence-modeling capabilities of Transformers. Diffusion-based approaches \cite{zhang2024motiondiffuse,zhang2023remodiffuse,tevet2022human} instead learn to progressively denoise motion representations, achieving strong generation quality and diversity. Other methods employ hierarchical representations, masked modeling, or retrieval augmentation to improve semantic alignment and motion fidelity \cite{li2026pressure2motion, ahmed2026k, xu2026vimorag}. Despite this progress, most methods encode the input and intermediate states using transient features, without explicitly preserving reusable knowledge throughout generation. A few recent approaches \cite{zhang2024infinimotion, kalakonda2025morag, li2026remomask} introduce memory through exemplar retrieval or long-context retention; however, they typically model only a single aspect of motion and fail to coordinate semantic intent, dynamic evolution, and generation history across scales.

We argue that effective memory for text-to-motion generation should jointly determine what motion to generate, how it should evolve, and how it remains consistent with the generated history. To this end, we organize semantic knowledge, motion dynamics, and causal context into a collaborative multi-scale memory system, enabling complementary guidance throughout generation.

Specifically, we introduce MMotion, a collaborative multi-scale memory framework that unifies scale-conditioned semantic retrieval, latent dynamics modeling, and causal working memory. Its cross-scale semantic memory retrieves scale- and position-conditioned motion priors, providing global structural guidance at coarse scales and fine-grained details at finer scales. Its latent dynamics memory models first- and second-order state changes to capture how motion evolves and to encourage plausible transitions. Its working memory summarizes previously generated states, preserving consistency between the current prediction and its generation history. We further design a collaborative fusion module that adaptively integrates the three memories and applies bounded residual corrections to the backbone features. Rather than optimizing each memory independently, MMotion jointly supervises semantic alignment, latent transitions, historical consistency, and inter-memory collaboration, allowing the three components to provide complementary instead of redundant guidance.

Moreover, to effectively inject memory without disrupting the pretrained generative prior, we devise a progressive training strategy that gradually activates memory fusion and lightweight adaptation. This design makes MMotion readily applicable to both diffusion-based and next scale autoregressive models. Extensive experiments on standard benchmarks and multiple representative backbones demonstrate consistent improvements over strong baselines in text–motion alignment, motion quality, diversity, and temporal coherence.  Our contributions can be summarized as follows:

\begin{itemize}
    \item  We propose MMotion, a collaborative multi-scale memory framework that formulates text-to-motion generation as the coordinated modeling of semantic intent, motion evolution, and generation history.
    \item We introduce three complementary memory mechanisms—cross-scale semantic memory, latent dynamics memory, and causal working memory—together with a joint fusion and optimization scheme for coherent memory-guided generation.
    \item  We develop an architecture-agnostic training strategy that enables MMotion to augment both diffusion-based and next scale autoregressive models, yielding consistent improvements across standard benchmarks and diverse backbones
\end{itemize}

\section{Related Work}
\label{sec:related}

\noindent\textbf{Diffusion-based text-to-motion generation.} Diffusion models have emerged as a prominent paradigm for text-conditioned motion synthesis. MotionDiffuse~\cite{zhang2024motiondiffuse} introduces fine-grained text conditioning into the diffusion process, while MDM~\cite{tevet2022human} incorporates geometric objectives to improve motion quality. MLD~\cite{chen2023executing} performs diffusion in a compact latent space, and ReMoDiffuse~\cite{zhang2023remodiffuse} augments denoising with retrieved text--motion exemplars. Recent studies have further explored efficient sampling, compositional generation, and structured motion representations. MotionLCM~\cite{dai2024motionlcm} accelerates synthesis through latent consistency modeling, EnergyMoGen~\cite{zhang2025energymogen} introduces energy-based composition in latent diffusion, and SALAD~\cite{hong2025salad} explicitly models interactions among skeletal joints, temporal frames, and textual tokens. Rethinking Diffusion~\cite{meng2025rethinking} revisits continuous motion representations and combines diffusion with masked autoregressive generation, while MotionHiFlow~\cite{li2026motionhiflow} adopts hierarchical flow matching for coarse-to-fine synthesis. These approaches primarily advance the generative process, representation, or conditioning mechanism. In contrast, MMotion provides complementary memory guidance and can be directly integrated into diffusion-based backbones.

\noindent\textbf{Autoregressive text-to-motion generation.} Autoregressive methods typically discretize motion sequences and formulate text-to-motion synthesis as conditional token prediction. TM2T~\cite{guo2022tm2t} employs vector quantization and sequence-to-sequence modeling to translate between text and motion tokens. T2M-GPT~\cite{zhang2023generating} combines a motion VQ-VAE with a GPT-style Transformer, while MotionGPT~\cite{jiang2023motiongpt} unifies language and motion within a shared token space. Subsequent methods improve motion tokenization and sequence modeling through residual quantization, masked prediction, and hierarchical decoding~\cite{guo2024momask,pinyoanuntapong2024mmm}. More recently, ScaMo~\cite{lu2025scamo} investigates scaling laws for autoregressive motion generation using a finite-scalar-quantized tokenizer and a text-prefix Transformer. MOGO~\cite{fu2026mogo} introduces hierarchical residual motion tokens for efficient streaming generation, and MoScale~\cite{zheng2026next} replaces conventional next-token prediction with coarse-to-fine next-scale autoregression. Unlike these methods, which encode generation context implicitly in token prefixes and hidden states, MMotion explicitly maintains semantic, dynamic, and causal memories. It therefore complements next scale autoregressive modeling without altering its underlying token-prediction formulation.

\noindent\textbf{Memory for text-to-motion generation.} Memory mechanisms have been introduced to provide motion generators with information beyond their instantaneous representations. ReMoDiffuse~\cite{zhang2023remodiffuse} retrieves relevant text--motion pairs from an external database and incorporates their motion features into diffusion-based denoising. MoRAG~\cite{kalakonda2025morag} extends retrieval-augmented generation paradigm through body-part-aware retrieval and multi-source motion fusion. InfiniMotion~\cite{zhang2024infinimotion} retains temporal information to support arbitrarily long motion generation. More recently, ReMoMask~\cite{li2026remomask} combines cross-modal retrieval with masked motion modeling and semantic spatio-temporal fusion. Existing approaches mainly use memory to retrieve motion exemplars or retain long-range context. MMotion instead unifies scale-conditioned semantic retrieval, latent dynamics modeling, and causal working memory, jointly optimizing their interactions to coordinate what motion to generate, how it evolves, and how it remains consistent with its history.
\subsection{Overview}
\label{sec:overview}

Given a text description $y$, text-to-motion generation aims to produce a motion sequence $\mathbf{x}=\{\mathbf{x}_{t}\}_{t=1}^{T}$ that matches the description and remains coherent over time. Most existing methods generate motion using only the text condition and the hidden states of the current sample. They cannot explicitly reuse motion patterns learned from other samples, reason about how the current motion should evolve, or retain a structured record of what has already been generated. Consequently, they may omit important action details, produce unnatural transitions, or gradually deviate from the intended motion.

\begin{figure*}[t]
    \centering
    \includegraphics[width=\linewidth]{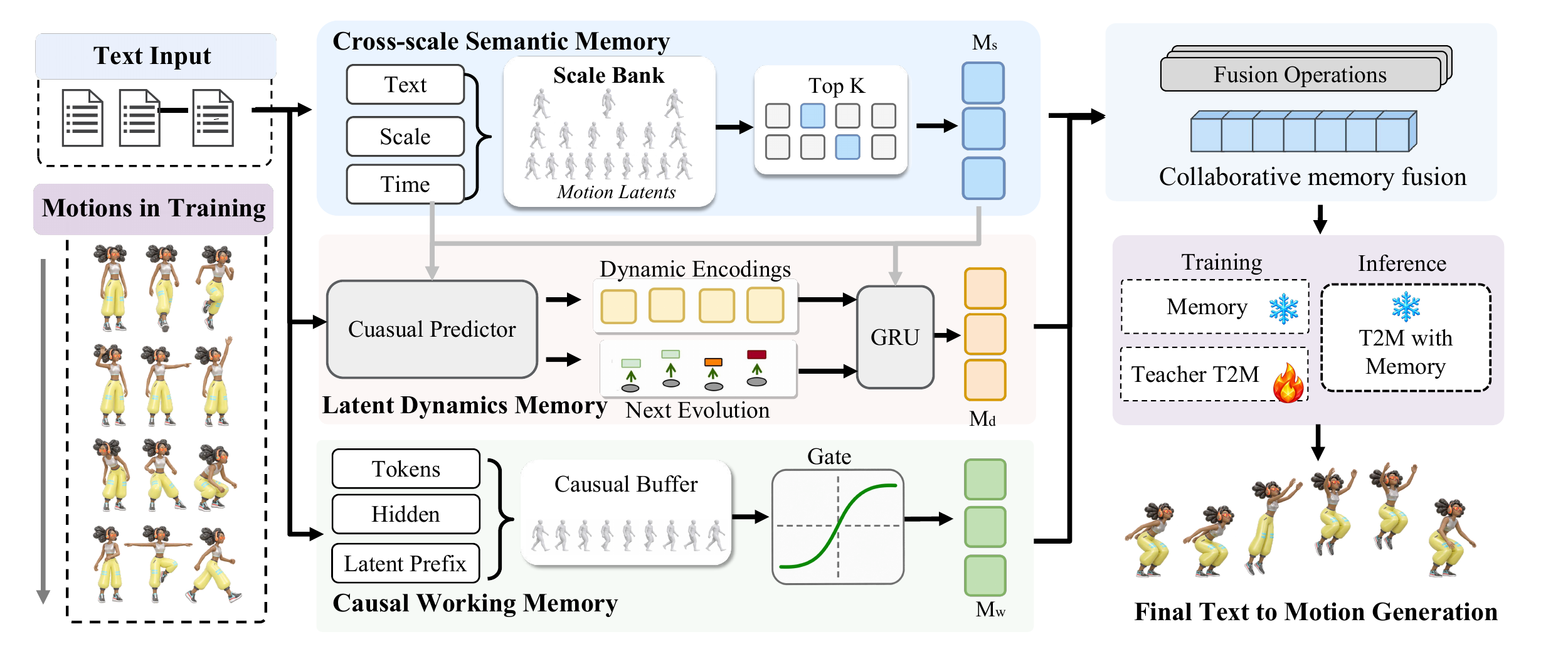}
     \vspace{-0.25in}
    \caption{\textbf{Overview of the MMotion pipeline.} In our work, cross-scale semantic, latent dynamics, and causal working memories jointly refine backbone features through a bounded residual correction, enabling coherent text-to-motion generation while preserving the pretrained model's generation pathway.}
    \label{fig:mmotion_pipeline}
    \vspace{-0.2in}
\end{figure*}

We propose \textbf{MMotion}, a unified memory-augmented framework for next-scale autoregressive and diffusion-based text-to-motion models. As shown in Fig.~\ref{fig:mmotion_pipeline}, MMotion contains three complementary modules. First, a cross-scale semantic memory retrieves motion patterns that match the text, generation scale, and position within the generation process. Second, a latent dynamics memory estimates how the current motion state should evolve. Third, a causal working memory summarizes the motion generated thus far. Together, these modules answer three questions: \emph{what} motion should be generated, \emph{how} it should evolve, and \emph{what} has already been generated.

The three memory outputs are fused to produce a small residual correction to the backbone feature. This design allows MMotion to improve the current prediction while preserving the knowledge and generation behavior of the pretrained model.

Let $\mathbf{e}$ denote the text embedding, $\mathbf{h}_{s,n}$ the backbone hidden state, $\mathbf{z}_{s,n}$ the motion latent, and $p_{s,n}$ the normalized position of generation step $n$ within scale $s$. For next-scale autoregressive models, $s$ follows the model's native coarse-to-fine scale hierarchy, and $p_{s,n}$ indicates the position within the corresponding within-scale denoising process. For diffusion models, we partition the $N$ reverse-denoising steps chronologically into coarse-to-fine stages. In our implementation, three scales are defined by the normalized stage endpoints $s\in\{0.3,0.6,1.0\}$, with $p_{s,n}$ denoting the relative position of step $n$ within each stage. In both cases, the refined feature is passed to the original prediction head, leaving the native sampling procedure unchanged.

We next introduce the three memory modules, their collaborative fusion, and the training strategy.

\subsection{Semantic, Dynamic, and Causal Memories}
\label{sec:memory}

\noindent \textbf{Cross-scale semantic memory.}
Motion descriptions contain information at different levels of detail. Coarse-scale information specifies the overall action, such as walking or sitting, whereas fine-scale information describes local details, such as arm movements or body orientation. A single shared representation may not capture both levels effectively. We therefore build a separate semantic memory bank for each generation scale:
\begin{equation}
    \mathcal{B}^{\mathrm{sem}}_{s}
    =
    \left\{
        \left(
            \mathbf{k}^{\mathrm{sem}}_{s,i},
            \mathbf{v}^{\mathrm{sem}}_{s,i}
        \right)
    \right\}_{i=1}^{N_s},
    \label{eq:semantic_bank}
\end{equation}
where $\mathbf{k}^{\mathrm{sem}}_{s,i}$ is a retrieval key and $\mathbf{v}^{\mathrm{sem}}_{s,i}$ stores the corresponding motion feature.

At each generation step, we form a query from the text embedding, the current scale, and the relative within-scale position:
\begin{equation}
    \mathbf{q}^{\mathrm{sem}}_{s,n}
    =
    \mathcal{Q}_{\mathrm{sem}}
    \left(
        \mathbf{e},s,p_{s,n}
    \right).
    \label{eq:semantic_query}
\end{equation}
We retrieve the top-$K$ entries whose keys are most similar to the query. Let $\mathcal{I}_{s,n}$ denote their index set. The retrieval weight of entry $i$ is
\begin{equation}
    w^{\mathrm{sem}}_{s,n,i}
    =
    \underset{i\in\mathcal{I}_{s,n}}
    {\operatorname{softmax}}
    \left(
        \frac{
            \operatorname{cos}
            \left(
                \mathbf{q}^{\mathrm{sem}}_{s,n},
                \mathbf{k}^{\mathrm{sem}}_{s,i}
            \right)
        }{\tau}
    \right),
    \label{eq:semantic_weight}
\end{equation}
where $\tau$ is a temperature parameter. The retrieved semantic feature is then
\begin{equation}
    \mathbf{m}^{\mathrm{sem}}_{s,n}
    =
    \sum_{i\in\mathcal{I}_{s,n}}
    w^{\mathrm{sem}}_{s,n,i}
    \mathbf{v}^{\mathrm{sem}}_{s,i}.
    \label{eq:semantic_readout}
\end{equation}
The semantic memory therefore performs the mapping
\begin{equation}
    (\mathbf{e},s,p_{s,n})
    \longmapsto
    \mathbf{m}^{\mathrm{sem}}_{s,n}.
\end{equation}
It provides global action information at coarse scales and local motion details at fine scales. Because its query does not depend on a specific generation mechanism, the same semantic memory can be used for both next-scale autoregressive and diffusion models.

\noindent \textbf{Latent dynamics memory.}
The semantic memory identifies relevant motion content, but it does not specify how the current motion should change. For example, knowing that a person is walking does not determine whether the next state should continue a step, initiate a turn, or slow down. We therefore introduce a latent dynamics memory to model the evolution of the current motion state.

We first compute first- and second-order differences between causally available latent states:
\begin{align}
    \mathbf{v}_{s,n}
    &=
    \mathbf{z}_{s,n}
    -
    \mathbf{z}_{s,n-1}, \\
    \mathbf{a}_{s,n}
    &=
    \mathbf{v}_{s,n}
    -
    \mathbf{v}_{s,n-1}.
    \label{eq:latent_differences}
\end{align}
These differences describe changes in latent space and should not be interpreted as physical joint velocity or acceleration. For next-scale autoregressive models, they are computed from latent states available along the within-scale denoising trajectory; for diffusion models, they are computed from preceding states along the reverse-denoising trajectory.

A causal predictor uses the current latent state, its recent change, and the generation conditions to predict the subsequent latent changes:
\begin{equation}
    \begin{aligned}
    \big(
        \mathbf{d}_{s,n},
        \widehat{\mathbf{v}}_{s,n+1},
        \widehat{\mathbf{a}}_{s,n+1},
        c^{\mathrm{dyn}}_{s,n}
    \big)
    =
    \mathcal{D}
    \big(
        \mathbf{z}_{s,n},
        \mathbf{v}_{s,n},
        \mathbf{e},
        s,p_{s,n}
    \big).
    \end{aligned}
    \label{eq:causal_predictor}
\end{equation}
Here, $\mathbf{d}_{s,n}$ encodes the current dynamics, $\widehat{\mathbf{v}}_{s,n+1}$ and $\widehat{\mathbf{a}}_{s,n+1}$ predict the subsequent latent changes, and $c^{\mathrm{dyn}}_{s,n}$ measures prediction confidence.

The dynamics encoding is used to retrieve similar transition patterns:
\begin{equation}
    \mathbf{m}^{\mathrm{dyn}}_{s,n}
    =
    \mathcal{M}_{\mathrm{dyn}}
    \left(
        \mathbf{d}_{s,n},
        s,p_{s,n},
        \mathbf{m}^{\mathrm{sem}}_{s,n}
    \right).
    \label{eq:dynamics_retrieval}
\end{equation}
Importantly, retrieval is also conditioned on the semantic feature. This prevents the model from selecting a transition that is smooth but inconsistent with the described action.

We then combine the retrieved transition with the predicted latent changes:
\begin{equation}
    \begin{aligned}
    \mathbf{q}^{\mathrm{dyn}}_{s,n}
    =
    \mathcal{F}_{\mathrm{dyn}}
    \big(
        \mathbf{m}^{\mathrm{sem}}_{s,n},
        \mathbf{m}^{\mathrm{dyn}}_{s,n},
        \mathbf{d}_{s,n},
        \widehat{\mathbf{v}}_{s,n+1},
        \widehat{\mathbf{a}}_{s,n+1},
        \bar c^{\mathrm{dyn}}_{s,n}
    \big),
    \end{aligned}
    \label{eq:dynamics_fusion}
\end{equation}
where $\bar c^{\mathrm{dyn}}_{s,n}$ is the calibrated confidence score.

A GRU propagates this dynamics information from coarse to fine scales:
\begin{equation}
    \mathbf{u}^{\mathrm{dyn}}_{s,n}
    =
    \operatorname{GRU}
    \left(
        \mathbf{q}^{\mathrm{dyn}}_{s,n},
        \mathbf{u}^{\mathrm{dyn}}_{s^{-},n}
    \right),
    \label{eq:dynamics_gru}
\end{equation}
where $s^{-}$ denotes the preceding scale. The dynamics module can thus be summarized as
\begin{equation}
    \left(
        \mathbf{z}_{s,\leq n},
        \mathbf{e},
        s,p_{s,n},
        \mathbf{m}^{\mathrm{sem}}_{s,n}
    \right)
    \longmapsto
    \mathbf{u}^{\mathrm{dyn}}_{s,n}.
\end{equation}

\noindent \textbf{Causal working memory.}
The semantic and dynamics memories store motion patterns shared across the dataset. However, they do not explicitly record what has already been generated for the current sample. This information is important because each new prediction should remain consistent with previous outputs. We therefore introduce a sample-specific causal working memory.

At step $(s,n)$, the working-memory buffer is
\begin{equation}
    \mathcal{C}_{s,n}
    =
    \left\{
        \mathbf{o}_{\prec(s,n)},
        \mathbf{h}_{\prec(s,n)},
        \mathbf{z}_{\prec(s,n)}
    \right\},
    \label{eq:causal_buffer}
\end{equation}
where $\prec(s,n)$ denotes all generation steps that causally precede $(s,n)$, and $\mathbf{o}$, $\mathbf{h}$, and $\mathbf{z}$ denote the corresponding generator outputs, hidden states, and latent states, respectively. For next-scale autoregressive models, the buffer contains outputs from completed coarser scales and preceding steps of the current within-scale denoising process. For diffusion models, it contains only states produced at preceding reverse-denoising steps.

A causal reader summarizes the buffer:
\begin{equation}
    \mathbf{m}^{\mathrm{work}}_{s,n}
    =
    \mathcal{M}_{\mathrm{work}}
    \left(
        \mathcal{C}_{s,n}
    \right).
    \label{eq:working_readout}
\end{equation}
Not all past information is equally useful at every step. We therefore use a gate to control the contribution of working memory:
\begin{equation}
    \begin{aligned}
    \mathbf{g}^{\mathrm{work}}_{s,n}
    =
    \sigma
    \Big(
        \mathcal{F}_{\mathrm{work}}
        \big(
            \mathbf{h}_{s,n},
            \mathbf{m}^{\mathrm{sem}}_{s,n},
            \mathbf{u}^{\mathrm{dyn}}_{s,n},
            \mathbf{m}^{\mathrm{work}}_{s,n}
        \big)
    \Big).
    \end{aligned}
    \label{eq:working_gate}
\end{equation}
The gated working-memory feature is
\begin{equation}
    \widetilde{\mathbf{m}}^{\mathrm{work}}_{s,n}
    =
    \mathbf{g}^{\mathrm{work}}_{s,n}
    \odot
    \mathbf{m}^{\mathrm{work}}_{s,n}.
    \label{eq:gated_working_memory}
\end{equation}

The buffer is reset for each sample. During inference, it stores only causally available information and never accesses outputs or latent states from future generation steps.

\subsection{Memory Collaboration and Joint Training}
\label{sec:joint_training}

\noindent \textbf{Collaborative memory fusion.}
The three memories play distinct but interconnected roles. Semantic memory determines what motion is relevant, and its output guides dynamics memory in selecting an appropriate transition. The semantic and dynamics features then help determine which parts of the generation history are useful. This sequential design allows the three memories to complement one another.

We fuse their final outputs as:
\begin{equation}
    \begin{aligned}
    \mathbf{j}_{s,n}
    =
    \mathcal{F}_{\mathrm{joint}}
    \big(
        \mathbf{m}^{\mathrm{sem}}_{s,n},
        \mathbf{u}^{\mathrm{dyn}}_{s,n},
        \widetilde{\mathbf{m}}^{\mathrm{work}}_{s,n},
        s,p_{s,n}
    \big).
    \end{aligned}
    \label{eq:joint_fusion}
\end{equation}
A residual head converts the fused feature into a candidate correction:
\begin{equation}
    \widehat{\Delta\mathbf{h}}_{s,n}
    =
    \rho_{s,n}
    \tanh
    \left(
        W_o\mathbf{j}_{s,n}
    \right),
    \label{eq:raw_residual}
\end{equation}
where $\rho_{s,n}$ controls how strongly the memories modify the backbone feature.

To avoid altering the pretrained representation too aggressively, we limit the correction according to the norm of the original hidden state:
\begin{equation}
    r_{s,n}
    =
    \tau_{\max}
    \left\|
        \mathbf{h}_{s,n}
    \right\|_2.
    \label{eq:residual_radius}
\end{equation}
The bounded residual is
\begin{equation}
    \Delta\mathbf{h}_{s,n}
    =
    \widehat{\Delta\mathbf{h}}_{s,n}
    \min
    \left(
        1,
        \frac{r_{s,n}}
        {
            \|
            \widehat{\Delta\mathbf{h}}_{s,n}
            \|_2+\epsilon_r
        }
    \right),
    \label{eq:bounded_residual}
\end{equation}
and the refined feature is
\begin{equation}
    \mathbf{h}'_{s,n}
    =
    \mathbf{h}_{s,n}
    +
    \Delta\mathbf{h}_{s,n}.
    \label{eq:corrected_feature}
\end{equation}
We set $\tau_{\max}=0.05$, limiting the correction norm to at most $5\%$ of the original hidden-state norm. This constraint preserves the pretrained generation pathway while allowing the memories to make targeted refinements.

\noindent \textbf{Trust-region distillation.}
Bounding the feature correction does not guarantee that the output distribution remains close to that of the pretrained model. We therefore freeze the original generator and use it as a teacher.

Let $\pi_{\mathrm{base}}^{T}$ and $\pi_{\mathrm{mem}}^{T}$ denote the softened predictive distributions of the original and memory-augmented models, respectively. For diffusion-based predictions, including the within-scale denoising process of next-scale autoregressive models, these correspond to the reverse-transition distributions induced by the original prediction head. We define
\begin{equation}
    \mathcal{L}_{\mathrm{trust}}
    =
    \max
    \left(
        0,\,
        D_{\mathrm{KL}}
        \left(
            \pi_{\mathrm{base}}^{T}
            \,\Vert\,
            \pi_{\mathrm{mem}}^{T}
        \right)
        -
        \epsilon
    \right),
    \label{eq:trust_loss}
\end{equation}
where $T=2$ is the softening temperature and $\epsilon$ specifies the allowed deviation from the teacher. This loss is active only when the divergence exceeds $\epsilon$, giving the memory modules room to improve the prediction without moving too far from the pretrained model.

We also use teacher uncertainty and calibrated dynamics confidence to adjust $\rho_{s,n}$. When the retrieved information is unreliable, MMotion applies a smaller correction.

\noindent \textbf{Training objective.}
We retain the native training objective of each backbone:
\begin{equation}
    \mathcal{L}_{\mathrm{gen}}
    =
    \begin{cases}
        \mathcal{L}_{\mathrm{NSAR}},
        & \text{next-scale autoregressive generation},\\
        \mathcal{L}_{\mathrm{diff}},
        & \text{diffusion generation},
    \end{cases}
    \label{eq:generation_loss}
\end{equation}
where $\mathcal{L}_{\mathrm{NSAR}}$ is the original next-scale autoregressive objective, including its within-scale denoising loss, and $\mathcal{L}_{\mathrm{diff}}$ is the original noise-, velocity-, or data-prediction objective of the diffusion model.

The complete training objective is
\begin{equation}
    \begin{aligned}
    \mathcal{L}
    ={}&
    \mathcal{L}_{\mathrm{gen}}
    +
    \lambda_{\mathrm{trust}}
    \mathcal{L}_{\mathrm{trust}}
    +
    \lambda_{\mathrm{sem}}
    \mathcal{L}_{\mathrm{sem}}
    \\ &+
    \lambda_{\mathrm{dyn}}
    \mathcal{L}_{\mathrm{dyn}}
    +
    \lambda_{\mathrm{work}}
    \mathcal{L}_{\mathrm{work}}
    +
    \lambda_{\mathrm{collab}}
    \mathcal{L}_{\mathrm{collab}}.
    \end{aligned}
    \label{eq:total_loss}
\end{equation}
Here, $\mathcal{L}_{\mathrm{sem}}$ aligns semantic queries with scale-specific motion features. $\mathcal{L}_{\mathrm{dyn}}$ supervises latent-change prediction and confidence estimation. $\mathcal{L}_{\mathrm{work}}$ encourages the compressed working memory to preserve information from the causally available generation history. Finally, $\mathcal{L}_{\mathrm{collab}}$ encourages the three memories to produce a consistent residual correction.

\noindent \textbf{Progressive training.}
Directly training all newly introduced modules together may disrupt the pretrained generator before the memories become reliable. We therefore use a three-phase training strategy.

In Phase A, the backbone is frozen and the residual correction is disabled. We train semantic retrieval, dynamics prediction, and working-memory encoding using their corresponding objectives. This phase allows each memory to learn its basic function without affecting the generator.

In Phase B, we gradually activate the residual correction and introduce teacher supervision and the trust-region loss. The three memories and the collaborative fusion layers are then optimized jointly, while the backbone remains frozen.

In Phase C, we activate low-rank adapters in the final layers of the generator and train them with a smaller learning rate. This allows the output layers to adapt to the memory-enhanced features after the memory modules have stabilized. We also enable retrieval-weight recalibration and margin convergence in this phase to improve late-stage training stability.

At inference, the teacher branch is removed.
\begin{table*}[t]
    \centering
    \caption{\textbf{Quantitative comparison on the HumanML3D benchmark.}
    $\uparrow$ indicates higher is better, $\downarrow$ indicates lower is better, and
    $\rightarrow$ indicates closer to the real-motion value is better.}
     \vspace{-0.15in}
    \label{tab:main_results}
    \setlength{\tabcolsep}{5pt}
    \renewcommand{\arraystretch}{1.10}
    \small
    \begin{tabular*}{\textwidth}{@{\extracolsep{\fill}}lcccccc@{}}
        \toprule
        \textbf{Method}
        & \multicolumn{3}{c}{\textbf{R-Precision} $\uparrow$}
        & \textbf{FID} $\downarrow$
        & \textbf{MM-Dist} $\downarrow$
        & \textbf{Diversity} $\rightarrow$ \\
        \cmidrule(lr){2-4}
        & \textbf{Top-1}
        & \textbf{Top-2}
        & \textbf{Top-3}
        & & & \\
        \midrule

        Real Motion
        & 0.511 & 0.703 & 0.797
        & 0.002 & 2.974 & 9.503 \\
        \midrule

        MotionDiffuse~\cite{zhang2024motiondiffuse}
        & 0.491 & 0.681 & 0.782
        & 0.630 & 3.113 & 9.410 \\

        ReMoDiffuse~\cite{zhang2023remodiffuse}
        & 0.510 & 0.698 & 0.795
        & 0.103 & 2.974 & 9.018 \\

        DiverseMotion~\cite{lou2023diversemotion}
        & 0.515 & 0.706 & 0.802
        & 0.072 & 2.941 & 9.683 \\

        MMM~\cite{pinyoanuntapong2024mmm}
        & 0.515 & 0.708 & 0.804
        & 0.089 & 2.926 & 9.577 \\

        MoMask~\cite{guo2024momask}
        & 0.521 & 0.713 & 0.807
        & 0.045 & 2.958 & -- \\

        BAMM~\cite{pinyoanuntapong2024bamm}
        & 0.525 & 0.720 & 0.814
        & 0.055 & 2.919 & 9.717 \\

        MoGenTS~\cite{yuan2024mogents}
        & 0.529 & 0.719 & 0.812
        & 0.033 & 2.867 & 9.570 \\

        Light-T2M~\cite{zeng2025light}
        & 0.511 & 0.699 & 0.795
        & 0.040 & 3.002 & -- \\

        IRG-MotionLLM~\cite{li2026irg}
        & 0.535 & 0.725 & 0.820
        & 0.242 & 2.785 & 9.900 \\

        EnergyMoGen~\cite{zhang2025energymogen}
        & 0.526 & 0.718 & 0.815
        & 0.176 & 2.931 & 9.500 \\

        MoScale ($S=4$)~\cite{zheng2026next}
        & 0.535 & 0.721 & 0.812
        & 0.049 & 2.867 & 9.510 \\

        MoScale ($S=18$)~\cite{zheng2026next}
        & 0.540 & 0.727 & 0.817
        & 0.046 & 2.830 & 9.525 \\

        MotionHiFlow~\cite{li2026motionhiflow}
        & 0.563 & 0.754 & 0.843
        & 0.032 & 2.691 & \textbf{9.504} \\
        \midrule

        \rowcolor{blue!8}
        \textbf{MoScale + MMotion ($S=18$, Ours)}
        & 0.554
        & 0.743
        & 0.827
        & 0.044
        & 2.794
        & 9.354 \\

        \rowcolor{blue!8}
        \textbf{MotionHiFlow + MMotion (Ours)}
        & \textbf{0.572}
        & \textbf{0.757}
        & \textbf{0.844}
        & \textbf{0.025}
        & \textbf{2.675}
        & 9.439 \\

        \bottomrule
    \end{tabular*}
\end{table*}

\begin{figure*}[t]
    \centering
    \includegraphics[width=\linewidth]{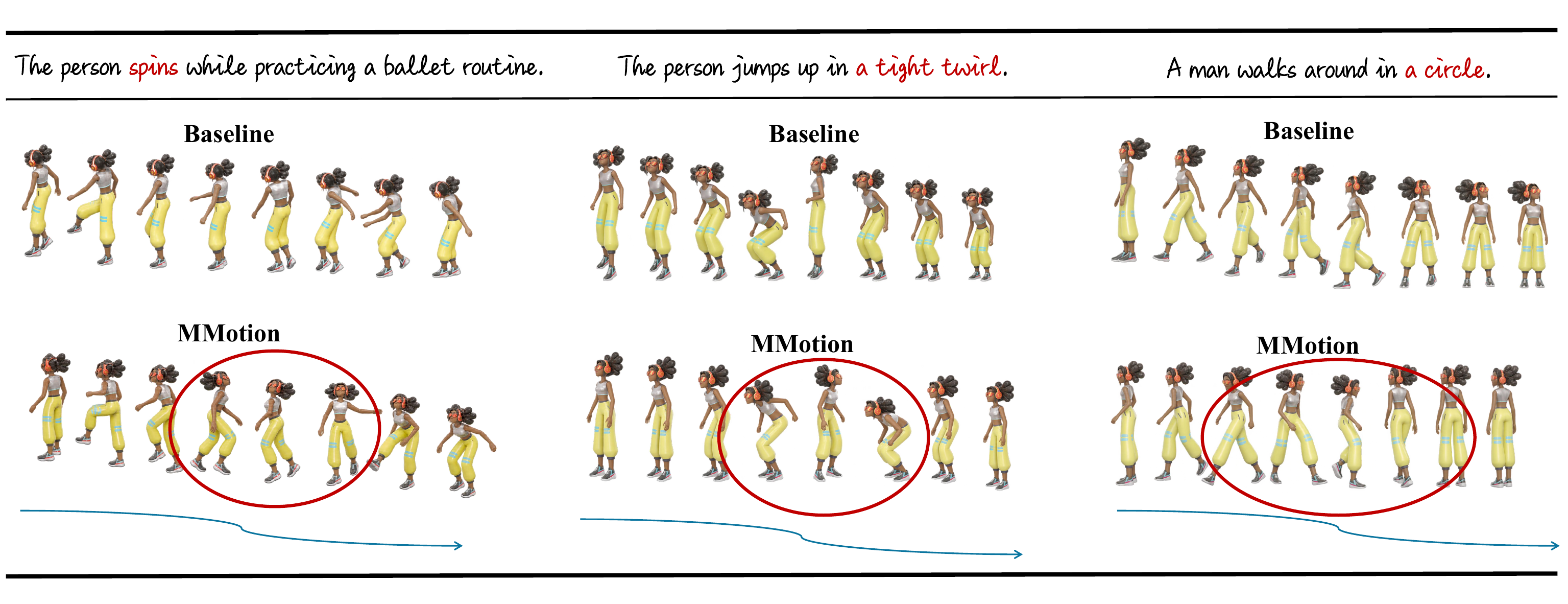}
    \vspace{-0.25in}
\caption{\textbf{Qualitative comparison on HumanML3D.} We compare MoScale equipped with MMotion against the original MoScale baseline. By leveraging semantic and dynamics memories, MMotion better distinguishes and realizes fine-grained motion concepts such as \emph{spins}, \emph{a tight twirl}, and \emph{a circle}. Its working memory further tracks generation progress, enabling the model to complete the circular trajectory rather than producing an incomplete or ambiguous motion. Red ellipses highlight the key motion patterns, and blue arrows indicate temporal progression.}
\label{fig:visul}
 \vspace{-0.2in}
\end{figure*}

\begin{table*}[t]
    \centering
    \caption{\textbf{Quantitative comparison on the KIT-ML benchmark.}
    $\uparrow$ indicates higher is better, $\downarrow$ indicates lower is better, and
    $\rightarrow$ indicates closer to the real-motion value is better.}
    \label{tab:kit_results}
     \vspace{-0.15in}
    \setlength{\tabcolsep}{5pt}
    \renewcommand{\arraystretch}{1.10}
    \small
    \begin{tabular*}{\textwidth}{@{\extracolsep{\fill}}lcccccc@{}}
        \toprule
        \textbf{Method}
        & \multicolumn{3}{c}{\textbf{R-Precision} $\uparrow$}
        & \textbf{FID} $\downarrow$
        & \textbf{MM-Dist} $\downarrow$
        & \textbf{Diversity} $\rightarrow$ \\
        \cmidrule(lr){2-4}
        & \textbf{Top-1}
        & \textbf{Top-2}
        & \textbf{Top-3}
        & & & \\
        \midrule

        Real Motion
        & 0.424 & 0.649 & 0.779
        & 0.031 & 2.788 & 11.080 \\
        \midrule

        MotionDiffuse~\cite{zhang2024motiondiffuse}
        & 0.417 & 0.621 & 0.739
        & 1.954 & 2.958 & 11.100 \\

        ReMoDiffuse~\cite{zhang2023remodiffuse}
        & 0.427 & 0.641 & 0.765
        & 0.155 & 2.814 & 10.800 \\

        DiverseMotion~\cite{lou2023diversemotion}
        & 0.416 & 0.637 & 0.760
        & 0.468 & 2.892 & 10.873 \\

        MMM~\cite{pinyoanuntapong2024mmm}
        & 0.404 & 0.621 & 0.744
        & 0.316 & 2.977 & 10.910 \\

        MoMask~\cite{guo2024momask}
        & 0.433 & 0.656 & 0.781
        & 0.204 & 2.779 & -- \\

        BAMM~\cite{pinyoanuntapong2024bamm}
        & 0.438 & 0.661 & 0.788
        & 0.183 & 2.723 & 11.008 \\

        MoGenTS~\cite{yuan2024mogents}
        & 0.445 & 0.671 & 0.797
        & 0.143 & 2.711 & 10.918 \\

        Light-T2M~\cite{zeng2025light}
        & 0.444 & 0.670 & 0.794
        & 0.161 & 2.746 & -- \\

        IRG-MotionLLM~\cite{li2026irg}
        & 0.445 & 0.681 & 0.781
        & 0.432 & 2.740 & 11.115 \\

        EnergyMoGen~\cite{zhang2025energymogen}
        & 0.436 & 0.651 & 0.772
        & 0.495 & 2.861 & \textbf{11.060} \\

        MoScale ($S=4$)~\cite{zheng2026next}
        & 0.441 & 0.662 & 0.783
        & 0.183 & 2.758 & 10.812 \\

        MoScale ($S=18$)~\cite{zheng2026next}
        & 0.442 & 0.671 & 0.791
        & 0.173 & 2.717 & 10.972 \\

        MotionHiFlow~\cite{li2026motionhiflow}
        & 0.482 & 0.704 & 0.825
        & \textbf{0.135} & 2.552 & 10.894 \\
        \midrule

        \rowcolor{blue!8}
        \textbf{MoScale + MMotion ($S=18$, Ours)}
        & 0.467
        & 0.640
        & 0.757
        & 0.455
        & 2.760
        & 11.226 \\

        \rowcolor{blue!8}
        \textbf{MotionHiFlow + MMotion (Ours)}
        & \textbf{0.528}
        & \textbf{0.737}
        & \textbf{0.835}
        & 0.137
        & \textbf{2.540}
        & 10.848 \\

        \bottomrule
    \end{tabular*}
\end{table*}

\section{Experiment}
\subsection{Experimental Details}
We evaluate our three-memory framework on HumanML3D \cite{guo2022generating} and KIT-ML \cite{plappert2016kit}, which represent human motion using 22 joints at 20\,FPS and 21 joints at 12.5\,FPS, respectively. We follow the data representations, preprocessing pipelines, and training, validation, and test splits adopted by each backbone. To prevent data leakage, the semantic and dynamics memories are constructed exclusively from the training set, whereas the working memory is maintained independently for each generated sequence and is not shared across test samples. 

We integrate our framework into two representative generation paradigms: the discrete multi-scale autoregressive model MoScale and the hierarchical flow-matching model MotionHiFlow. Both models are initialized from their corresponding pretrained checkpoints; we freeze the generative backbones and train only the memory-related modules for five epochs. Semantic and dynamics retrieval use the top eight candidates with a temperature of 0.06. Retrieved information is incorporated through bounded residual connections, together with auxiliary supervision and a trust-region constraint to preserve the pretrained generative prior. For MoScale, we use a batch size of 32 and a learning rate of \(5\times10^{-5}\). Generation performs \([1,2,5,10]\) sampling steps across four scales (\(S=18\)), with a sampling temperature of 1.0 and top-\(p=0.9\). For MotionHiFlow, we use a batch size of 16 and learning rates of \(3\times10^{-5}\) and \(1\times10^{-4}\) on HumanML3D and KIT-ML, respectively, while retaining its native sampling procedure. 

We report R-Precision (Top-1/2/3), Fr\'echet Inception Distance (FID), multimodal distance (MM-Dist), and Diversity. Higher R-Precision and lower FID and MM-Dist indicate better performance, whereas Diversity is assessed by its proximity to the corresponding real-motion value.

\subsection{Main Results}
\noindent \textbf{Results on HumanML3D.}
As shown in Table~\ref{tab:main_results}, MMotion consistently improves both backbones, demonstrating that the proposed memory framework generalizes across discrete autoregressive and continuous flow-matching paradigms. When integrated with MoScale, MMotion improves all R-Precision metrics while preserving competitive distributional fidelity, indicating that memory retrieval strengthens text--motion correspondence without substantially disrupting the pretrained motion prior. Its integration with MotionHiFlow yields the strongest overall performance, establishing new best results in semantic alignment, FID, and MM-Dist while maintaining diversity close to that of real motions. These complementary gains suggest that the three memory components address distinct limitations of parametric generation: semantic memory resolves textual ambiguity, dynamics memory supplies reusable motion patterns, and working memory preserves temporal consistency throughout generation. The simultaneous improvements in retrieval accuracy and distributional quality further indicate that MMotion does not merely bias generation toward easily recognizable actions, but instead produces motions that are both semantically faithful and realistic. We provide a qualitative analysis in Fig.~\ref{fig:visul}, with additional visualizations included in the supplementary material.
\begin{table*}[t]
    \centering
    \caption{\textbf{Ablation studies of MMotion on HumanML3D.}
    MMotion is instantiated with MoScale, and all variants are evaluated using the same
    checkpoint and fixed random seed. R-Precision is reported as a percentage.
    $\uparrow$ indicates higher is better, $\downarrow$ indicates lower is better, and
    $\rightarrow$ indicates closer to the real-motion value is better.}
      \vspace{-0.15in}
    \label{tab:mmotion_ablation}
    \setlength{\tabcolsep}{3.8pt}
    \renewcommand{\arraystretch}{1.08}
    \scriptsize
    \resizebox{\textwidth}{!}{
    \begin{tabular}{lccccccc}
        \toprule
        \textbf{Configuration}
        & \textbf{R@1} $\uparrow$
        & \textbf{R@2} $\uparrow$
        & \textbf{R@3} $\uparrow$
        & \textbf{FID} $\downarrow$
        & \textbf{MM-Dist} $\downarrow$
        & \textbf{Diversity} $\rightarrow$
        & \textbf{Multimodality} $\uparrow$ \\
        \midrule

        \multicolumn{8}{l}{\textit{(a) Memory-injection scales}} \\
        Coarse scales only
        & \textbf{55.409} & 74.203 & 82.478 & 0.04732 & 2.79769 & 9.37935 & 0.73822 \\
        Fine scales only
        & 55.259 & 74.116 & 82.565 & 0.04625 & 2.79924 & 9.33960 & \textbf{0.77295} \\
        \rowcolor{blue!8}
        \textbf{All scales (Full)}
        & \textbf{55.409} & \textbf{74.267} & \textbf{82.694}
        & \textbf{0.04440} & \textbf{2.79426} & 9.35375 & 0.74710 \\
        \midrule

        \multicolumn{8}{l}{\textit{(b) Number of retrieved entries}} \\
        $K=2$
        & 55.323 & \textbf{74.418} & 82.672 & 0.04878 & 2.79316 & 9.32286 & \textbf{0.74885} \\
        $K=4$
        & 55.129 & 74.267 & \textbf{82.909} & 0.04822 & 2.79664 & 9.35734 & 0.74291 \\
        \rowcolor{blue!8}
        $\boldsymbol{K=8}$ \textbf{(Full)}
        & \textbf{55.409} & 74.267 & 82.694 & \textbf{0.04440} & 2.79426 & 9.35375 & 0.74710 \\
        $K=16$
        & 55.043 & \textbf{74.418} & \textbf{82.909} & 0.04539 & \textbf{2.79113} & 9.38122 & 0.73409 \\
        \midrule

        \multicolumn{8}{l}{\textit{(c) Internal memory mechanisms}} \\
        Shuffled semantic associations
        & 55.065 & 73.297 & 82.198 & 0.04986 & 2.80428 & 9.38528 & \textbf{0.79311} \\
        Without dynamics-memory retrieval
        & 55.086 & 73.750 & 82.478 & 0.04912 & 2.80720 & 9.38364 & 0.75514 \\
        Mean aggregation for working memory
        & 55.108 & 74.116 & 82.608 & \textbf{0.04432} & 2.79770 & 9.36368 & 0.74535 \\
        Without working-memory correction
        & 55.302 & 74.052 & \textbf{82.866} & 0.04747 & \textbf{2.79383} & 9.37607 & 0.75249 \\
        \rowcolor{blue!8}
        \textbf{Complete MMotion}
        & \textbf{55.409} & \textbf{74.267} & 82.694
        & 0.04440 & 2.79426 & 9.35375 & 0.74710 \\
        \midrule

        \multicolumn{8}{l}{\textit{(d) Incremental memory components}} \\
        C0: All memory residuals disabled
        & 54.569 & 73.966 & 82.198 & 0.04588 & 2.80816 & 9.38111 & 0.77122 \\
        C1: Semantic memory
        & 55.043 & 74.052 & 82.069 & 0.04739 & 2.80303 & 9.36741 & 0.78481 \\
        C2: Semantic + dynamics memories
        & 55.323 & \textbf{74.547} & 82.586 & 0.04818 & 2.80159 & 9.37137 & \textbf{0.78887} \\
        \rowcolor{blue!8}
        \textbf{Full: Complete MMotion}
        & \textbf{55.409} & 74.267 & \textbf{82.694}
        & \textbf{0.04440} & \textbf{2.79426} & 9.35375 & 0.74710 \\
        \bottomrule
    \end{tabular}}
\end{table*}
\noindent \textbf{Results on KIT-ML.}
Table~\ref{tab:kit_results} further demonstrates the effectiveness of MMotion under the smaller and more heterogeneous KIT-ML benchmark. MotionHiFlow equipped with MMotion achieves the best R-Precision and MM-Dist, substantially strengthening text--motion alignment while retaining competitive FID and realistic diversity. This result is particularly informative because the memory modules are trained without modifying the pretrained backbone, showing that retrieved semantic and dynamics knowledge can compensate for limited supervision without sacrificing the underlying generative prior. The improvements are less uniform for MoScale: MMotion enhances Top-1 retrieval while Top-2/3 accuracy and MM-Dist do not consistently improve. This contrast suggests that memory cues are more naturally assimilated by the continuous, hierarchical refinement process of MotionHiFlow than by MoScale's discrete multi-scale decoding, where retrieved information may compete with early token decisions. Nevertheless, the strong and consistent gains obtained with MotionHiFlow across both benchmarks establish MMotion as an effective plug-in framework rather than an improvement tied to a particular dataset.

\subsection{Ablation Study}

In this paper, we conduct controlled ablations on HumanML3D to examine how the key design choices of MMotion affect its performance. All variants share the same checkpoint and random seed and differ only in the specified inference-time intervention.

\noindent \textbf{Where should memory be injected?}
We conduct controlled ablations on HumanML3D to examine how the key design choices of MMotion affect its performance. All variants share the same checkpoint and random seed and differ only in the specified inference-time intervention.
MMotion is designed to guide the entire generation process rather than a single stage. We therefore restrict memory injection to either coarse or fine scales in Table~\ref{tab:mmotion_ablation}(a). Both variants remain effective, but neither provides the balanced performance of all-scale injection, which yields the best FID, MM-Dist, and Top-2/3 R-Precision. This comparison supports applying memory throughout the generation hierarchy, such that retrieved knowledge can shape global motion structure at coarse scales and refine local dynamics at fine scales.

\noindent \textbf{How much information should be retrieved?}
Having established the value of all-scale injection, we next vary the number of retrieved entries in Table~\ref{tab:mmotion_ablation}(b). Retrieving too few entries may omit useful motion patterns, whereas retrieving too many may introduce less relevant information. Although different values favor individual metrics, $K=8$ achieves the best Top-1 R-Precision and FID, offering the most favorable balance between retrieval coverage and relevance. We therefore use $K=8$ as the default setting for MMotion.

\noindent \textbf{Does the internal memory design matter?}
We next determine whether MMotion benefits from meaningful memory operations rather than merely from additional residual features. As shown in Table~\ref{tab:mmotion_ablation}(c), shuffling semantic associations degrades R-Precision, FID, and MM-Dist, demonstrating the importance of retrieving correctly matched semantic knowledge. Disabling dynamics-memory retrieval likewise weakens alignment and distributional fidelity, confirming the contribution of reusable dynamics priors. Although simplified working-memory variants improve isolated metrics, the complete design provides a stronger overall trade-off, supporting the complementary roles of its internal mechanisms.

\noindent \textbf{How do the three memories work together?}
Finally, Table~\ref{tab:mmotion_ablation}(d) progressively activates the semantic, dynamics, and working memories to reveal how their contributions accumulate. Semantic memory first improves text--motion alignment, while adding dynamics memory further strengthens Top-1/2 R-Precision and MM-Dist. Incorporating working memory completes MMotion and yields the best Top-1/3 R-Precision, FID, and MM-Dist, suggesting that tracking the evolving generation helps reconcile retrieved knowledge with the motion already synthesized. Although not every metric improves monotonically, the complete framework achieves the most balanced semantic alignment and distributional fidelity.

\section{Conclusion}
This paper proposes MMotion, a model-agnostic three-memory framework that enhances text-driven motion generation through semantic, dynamics, and working memories. MMotion can be integrated into pretrained generators while keeping their backbones frozen. Experiments on HumanML3D and KIT-ML show consistent improvements across next scale autoregressive and flow-matching paradigms, particularly in text--motion alignment and motion quality. These results demonstrate the potential of explicit memory to improve motion generation efficiently without retraining the underlying generative model.

{
    \small
    \bibliographystyle{ieeenat_fullname}
    \bibliography{main}
}


\end{document}